\documentclass{article}

\usepackage{spconf,amsmath,graphicx}
\usepackage{todonotes}
\usepackage{soul}
\usepackage{hyperref}
\usepackage{xspace}
\usepackage{blindtext}

\definecolor{TodoColor}{rgb}{1,0.7,0.6}

\usepackage{xstring}
\usepackage{seqsplit}
\usepackage{placeins}

\newcommand{\capitalfirst}[1]{%
    \StrLeft{#1}{1}[\firstletter]%
    \StrGobbleLeft{#1}{1}[\restofword]%
    \MakeUppercase{\firstletter}\restofword%
}

\newcommand{\capitalizehyphenated}[1]{%
    \StrCut{#1}{-}{\firstpart}{\restpart}%
    \capitalfirst{\firstpart}%
    \IfStrEq{\restpart}{}{}{ \capitalizehyphenated{\restpart}}
}

\newcommand{\hfmodel}[1]{%
    \StrBehind{#1}{/}[\hfmodelshortname]%
    \href{https://huggingface.co/#1}{\capitalizehyphenated{\hfmodelshortname}}%
}

\usepackage[capitalise]{cleveref}
\crefname{figure}{Figure}{Figures}
\crefname{table}{Table}{Tables}
\crefname{appendix}{Appendix}{Appendices}

\usepackage{multirow}

\usepackage[table]{xcolor}
\definecolor{taskgreen}{RGB}{180, 220, 180}
\definecolor{taskblue}{RGB}{180, 210, 230}
\definecolor{taskorange}{RGB}{255, 210, 170}
\definecolor{taskpink}{RGB}{255, 185, 200}

\renewcommand{\paragraph}[1]{\medskip\noindent\textbf{#1}\quad}

\usepackage{booktabs}
\usepackage{tabularx}
\usepackage{arydshln}

\makeatletter
\def\adl@drawiv#1#2#3{%
  \hskip.5\tabcolsep
  \xleaders#3{#2.5\@tempdimb #1{1}#2.5\@tempdimb}%
    #2\z@ plus1fil minus1fil\relax
  \hskip.5\tabcolsep
}
\newcommand{\cdashlinelr}[1]{%
  \noalign{\vskip\aboverulesep
           \global\let\@dashdrawstore\adl@draw
           \global\let\adl@draw\adl@drawiv}%
  \cdashline{#1}%
  \noalign{\global\let\adl@draw\@dashdrawstore
           \vskip\belowrulesep}%
}
\makeatother

\usepackage{amssymb}
\usepackage{makecell}

\usepackage{orcidlink}
\usepackage{etoolbox}
\hypersetup{
  urlcolor     = blue, 
  linkcolor    = blue, 
}

\usepackage[table]{xcolor}

\usepackage{enumitem}
\setlist{noitemsep,left=0mm,topsep=0mm}
\usepackage{xfrac}
\usepackage{bbm}

\usepackage{booktabs}

\title{Controlling Backchannels in Streamable Full-duplex Models}
\name{\parbox{\dimexpr\textwidth-2\tabcolsep\relax}{\centering
Maike Z\"{u}fle\,\orcidlink{0009-0001-7238-7705}\textsuperscript{1},
Peter Pol\'{a}k\,\orcidlink{0000-0003-2332-7666}\textsuperscript{2,3},\\
Sefik Emre Eskimez\,\orcidlink{0000-0001-6259-5925}\textsuperscript{4},
Jan Niehues\,\orcidlink{0000-0002-4231-6543}\textsuperscript{1},
Peter Bell\,\orcidlink{0000-0002-9597-9615}\textsuperscript{5},
Ond\v{r}ej Klejch\,\orcidlink{0000-0001-5495-967X}\textsuperscript{5}
}}

\address{\parbox{\dimexpr\textwidth-2\tabcolsep\relax}{\centering
\textsuperscript{1}Karlsruhe Institute of Technology, Germany 
\textsuperscript{2}AppTek, Germany\\ \textsuperscript{3} Charles University, Czech Republic
\textsuperscript{4}Sesame AI, USA
\textsuperscript{5}University of Edinburgh, UK\\
\href{mailto:maike.zuefle@kit.edu}{\texttt{maike.zuefle@kit.edu}}
}}
\begin{document}
%
\maketitle
\begin{abstract}
Backchannels, brief acknowledgements like ``uh-huh'' produced while the other party may still be talking, are central to natural conversation, but full-duplex spoken dialogue models rarely model them explicitly. We introduce a lightweight backchannel head that predicts, from a full-duplex model's own hidden states, when a backchannel should begin. Once this probability crosses a tunable threshold, a backchannel is force-decoded. Attached to both a 7B (PersonaPlex) and a 1B (F-Actor) model, it generalizes across scale. Probing confirms the hidden states anticipate real human timing, and generation evaluation shows more frequent, better-timed backchannels. Human raters judge the resulting backchannels on par with real ones.

\end{abstract}

\section{Introduction}
\label{sec:intro}

Full-duplex spoken dialogue models~\cite{hu2025salmduplexefficientdirectduplex, défossez2024moshispeechtextfoundationmodel, roy2026personaplexvoicerolecontrol,zufle-etal-2026-f} can listen and speak simultaneously, making them well suited for generating backchannels, i.e., short acknowledgements, like ``uh-huh'', 
a listener produces while the other party may still be talking. 

Backchannel modelling received attention before the advent of full-duplex modelling, with earlier systems using explicit prediction heads to generate backchannels and improve rapport~\cite{lala2017attentive, ruede2017yeahrightuhhuhdeep}. Voice Activity Projection (VAP) models predict backchanneling frame-by-frame as an external classifier in a cascaded pipeline~\cite{inoue-etal-2025-yeah, inoue-etal-2026-multilingual}. Endpointing targets the more general problem of turn boundaries~\cite{udupa2026endpointanticipationlowlatencyspoken}. Complementary work studies the lexico-prosodic 
meaning of backchannels \cite{qian2026aligning}.

In full-duplex dialogue modelling, recent models produce backchannels as a byproduct of training on conversational data \cite{roy2026personaplexvoicerolecontrol, zufle-etal-2026-f}. Reinforcement learning can further shape this behaviour into more natural backchanneling \cite{ohashi2026multifacetedinteractivityalignmentfullduplex}. Though effective, this is expensive and yields an implicit policy with no controllable signal for when or why backchannels occur. Closer to explicit control, one line of work first pretrains a full-duplex model and then adds VAP heads for turn-taking \cite{rajaa2026dualturn}, while another probes a full-duplex model's representations for turn-taking without pretraining \cite{riera2026synchronization}.

Other existing evaluations of full-duplex backchanneling focus on whether models backchannel at all \cite{défossez2024moshispeechtextfoundationmodel,zufle-etal-2026-f}, and not whether those backchannels are appropriate.  Therefore, it is critical to evaluate backchannel placement against real human conversational timing, and to train a model that explicitly predicts when a backchannel should occur.

We address this gap with a lightweight, model-independent backchannel head. It predicts, at each timestep, the probability that a backchannel should begin, and when this probability crosses a threshold, we force-decode a backchannel token into the text stream, which in turn drives the corresponding audio output. This is in the spirit of frame-wise VAP-based backchannel predictors \cite{inoue-etal-2025-yeah}, but attached directly to a full-duplex model's own hidden states rather than run as an external classifier. We attach this head to two full-duplex models at different scales, the 7B PersonaPlex \cite{roy2026personaplexvoicerolecontrol} and the 1B F-Actor \cite{zufle-etal-2026-f}, 
to show that the method generalizes.
Since F-Actor does not support streaming inference, we additionally adapt it into a streaming model to enable this comparison.

We evaluate the resulting models along three complementary 
axes. First, we measure whether the model backchannels more often than the unmodified baseline. Second, we probe the models' hidden states to test whether they predict backchannel onset where real speakers 
actually  backchanneled in human conversations, 
using human-annotated timing as ground truth. Third, we run a human evaluation in which participants rate how appropriate the backchannels in generated dialogues sound, finding that they are rated on par with human backchannels.

Our contributions are as follows:
\begin{itemize}
    \item A model-independent method for controlling backchannel timing, which we show generalizes across two architectures and two model scales.
    \item A human-aligned evaluation of backchannel behaviour, combining a probing analysis against real human timing data with a human appropriateness study.
    \item A streaming version of F-Actor to validate our method on models of multiple sizes.
\end{itemize}
We release the code and models openly.\footnote{\href{https://github.com/MaikeZuefle/bcmore}{https://github.com/MaikeZuefle/bcmore}}

\section{Controllable Backchanneling}
\label{sec:head}
This section presents our proposed model-independent backchannel head.
Our design fulfils two objectives:
(1) At each frame, the model predicts whether the current point in the interlocutor's speech is an appropriate moment for a backchannel, i.e., a moment at which a human would produce one.
(2) The system must provide an independent control mechanism to adjust how frequently backchannels are emitted without degrading the appropriateness of their placement.

\subsection{Backchannel Head}
\label{ssec:head_arch}

To predict backchannel timing without altering the backbone's language modelling capabilities, we attach a lightweight MLP head directly to its hidden states. Let $\mathbf{h}_t \in \mathbb{R}^{d}$ denote the hidden state of the backbone (the temporal transformer) at timestep $t$. The head maps $\mathbf{h}_t$ to a probability via a two-layer MLP with a sigmoid output $\sigma(\cdot)$:
\begin{equation}
    p_t = \sigma(\mathbf{W}_2 \operatorname{GELU}(\mathbf{W}_1 \mathbf{h}_t + \mathbf{b}_1) + b_2),
\end{equation}
where $p_t \in [0, 1]$ is the predicted probability of a backchannel onset, i.e., the first frame of a backchannel, at $t+1$, and $b_2$ is initialized to the empirical log-odds of the class prior.

To handle extreme class imbalance (onsets account for~$\sim$1\% of conversational frames), the head is trained using Focal Loss~\cite{lin2017focal} evaluated only over eligible frames $\mathcal{T}_{\text{valid}}$ where the partner is speaking and the agent is silent. All frames during agent speech or mutual silence are excluded from $\mathcal{T}_{\text{valid}}$.
Let $y_t \in \{0, 1\}$ denote the frame-level target backchannel onset label at timestep $t+1$ and let $\tilde{p}_t = y_t p_t + (1 - y_t)(1 - p_t)$ denote the probability assigned to the target class using the predicted probabilities $p_t$. The Focal Loss $\mathcal{L}_{\text{head}}$ is then computed as:

\begin{equation}
    \mathcal{L}_{\text{head}} = -\frac{1}{|\mathcal{T}_{\text{valid}}|} \sum_{t \in \mathcal{T}_{\text{valid}}} \alpha_t (1 - \tilde{p}_t)^\gamma \log(\tilde{p}_t),
\end{equation}
where $\alpha_t = \alpha y_t + (1 - \alpha)(1 - y_t)$ balances class frequencies with hyperparameters $\alpha = 0.9$ and $\gamma = 2.0$.

\subsection{Thresholding and Conditioned Text Generation}
\label{ssec:inference}
During inference, we decouple placement detection from token emission via a threshold $\tau \in [0, 1]$: a backchannel is triggered whenever $p_t \ge \tau$, giving control over backchanneling frequency independently of placement quality.

Upon triggering at timestep $t$, we force-decode the special \texttt{[BC]} token into the model's text stream, replacing the  new-word marker \texttt{[EPAD]}. The model then continues autoregressive generation to produce the backchannel's lexical form (e.g., ``yeah'', ``uh-huh'', ``right''), conditioned on the injected token and surrounding context.

Critically, we mask the loss on the \texttt{[BC]} token itself, so the backbone never learns to emit it spontaneously, leaving initiation entirely to the thresholded head, while retaining standard cross-entropy loss on the subsequent transcript tokens that realize the backchannel's content.

\section{Experimental Setup}
\label{sec:exp}
\subsection{Models}
We evaluate our approach on two full-duplex models. We train \textbf{PersonaPlex (7B)}~\cite{roy2026personaplexvoicerolecontrol} with the backchannel mechanism from \cref{sec:head} on a single H100 80GB GPU for approximately 10 hours. The backbone is fine-tuned during training using LoRA~\cite{hu2021loralowrankadaptationlarge}, while the depth transformer is kept frozen.

In addition, we adapt \textbf{F-Actor (1B)}~\cite{zufle-etal-2026-f}. Backchanneling requires listening and speaking in real time, since backchannels overlap with the interlocutor's speech. F-Actor, however, does not support streaming inference, as it relies on NanoCodec~\cite{casanova2025nanocodec}. We therefore replace NanoCodec with the causal Mimi codec~\cite{défossez2024moshispeechtextfoundationmodel} and, following \cite{torgashov2026voxtream2}, add a depth transformer pretrained from \hfmodel{sesame/CSM-1B} (frozen) to model Mimi's codebooks. Speaker conditioning uses ECAPA-TDNN~\cite{desplanques2020ecapa} embeddings.
The temporal transformer backbone is \hfmodel{meta-llama/Llama-3.2-1B-Instruct} \cite{grattafiori2024llama3herdmodels}, with the BC head from \cref{ssec:head_arch} attached. Unlike PersonaPlex, we train it only to predict the system channel's inner monologue and Mimi codes. Training takes 13 hours on four A100 80GB GPUs.

Models with the backchannel head carry the suffix -BC.

\begin{figure*}
    \centering
    \includegraphics[width=0.9\linewidth]{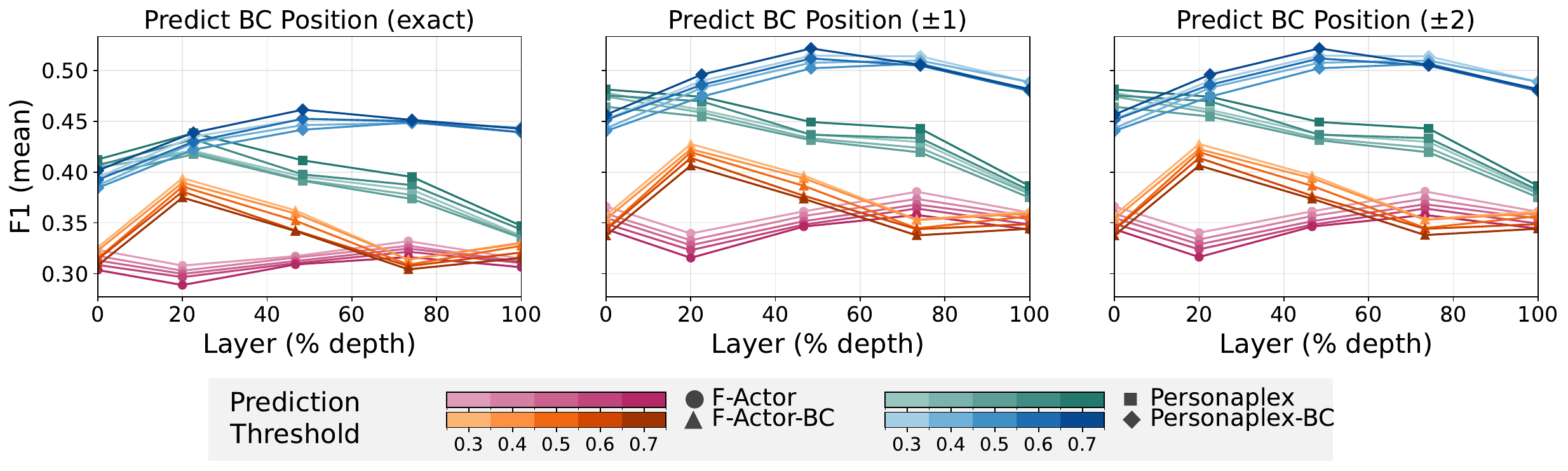}
    \caption{Correct backchannel timing F1 score per layer for predicting backchannel onset from frozen hidden states on TurnBench, for base and backchannel-head-trained (BC) models. A prediction only counts as correct if it lands at the exact annotated frame.}
    \label{fig:f1_probing}
\end{figure*}
\begin{figure}
    \centering
    \includegraphics[width=1.0\linewidth]{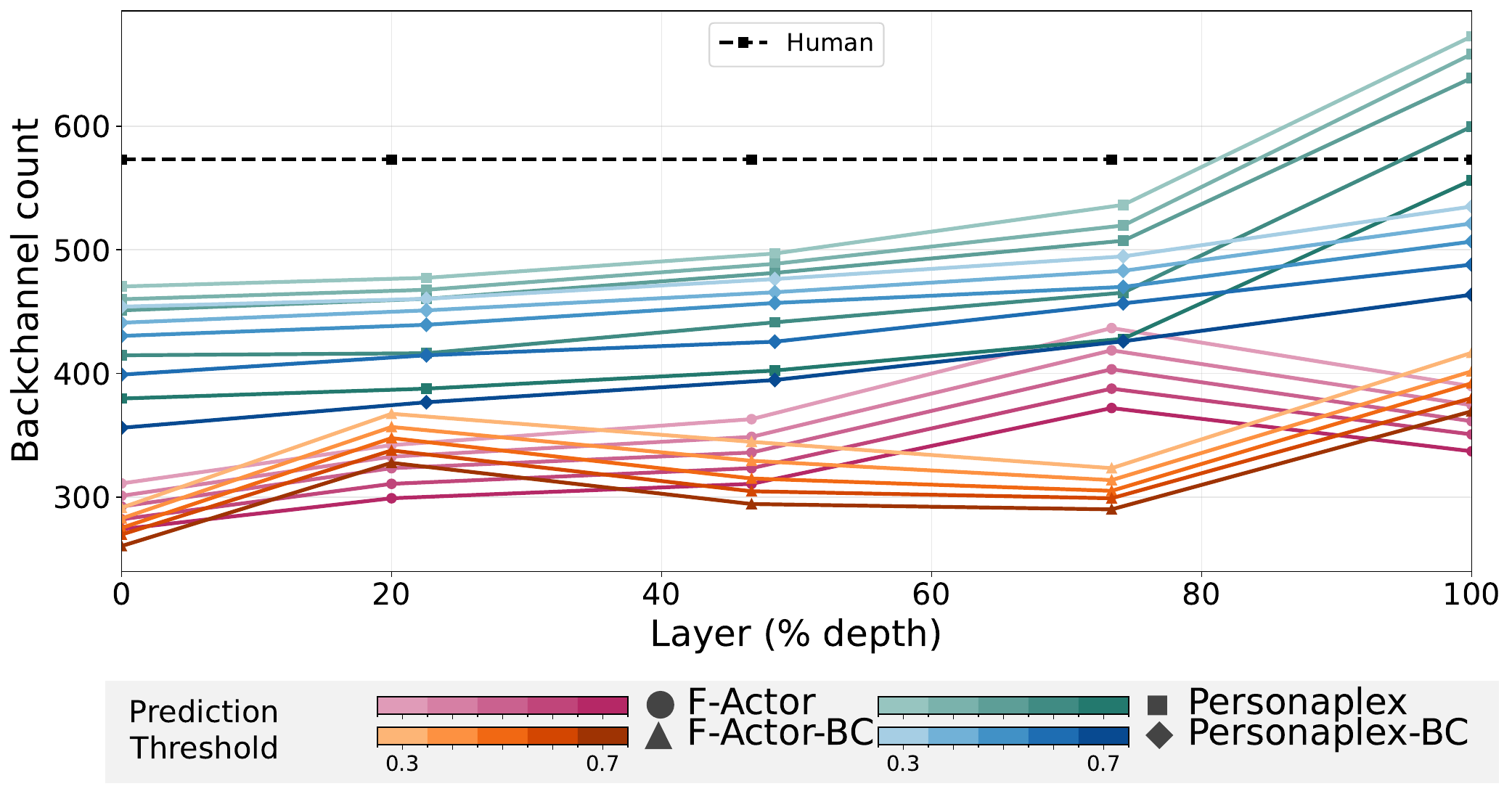}
    \caption{Predicted backchannel frequency per layer, for base and backchannel-head-trained (BC) models on TurnBench.}
    \label{fig:frequency}
\end{figure}

\subsection{Data}
\paragraph{Training.} We use Fisher~\cite{cieri2004fisher}, which contains around 2000h of English conversations, restoring the 8kHz audio into 24kHz with Sidon~\cite{nakata2026sidon}. Each conversation is transcribed with \hfmodel{nvidia/Parakeet-TDT-0.6B-v3}~\cite{sekoyan2025canary1bv2parakeettdt06bv3efficient}, and split into 90s chunks at the nearest utterance end. To condition the model to reliably produce backchannels upon receiving the \texttt{[BC]} token, we augment these dialogues by inserting additional backchannels from other occurrences at locations where the agent is silent for at least 1.5 seconds, and the interlocutor keeps speaking for at least 1s, constraints chosen after pilot experiments.

\paragraph{Evaluation.} We use TurnBench~\cite{jiang2026turnbench}, human-human conversations with backchannel annotations from three annotators per speaker (majority vote). We restrict to the twelve conversations where all three segmented the same turns: 137 minutes, 574 backchannels out of 1,888 total utterances. Each conversation is split into 60s windows (plus 10s prior context), snapped to the nearest silence gap and capped at 120s, and evaluated with leave-one-conversation-out cross-validation. Since TurnBench gives only utterance-level timestamps, we obtain word-level ones via \hfmodel{nvidia/Parakeet-TDT-0.6B-v3}~\cite{sekoyan2025canary1bv2parakeettdt06bv3efficient} on each turn's own audio and transcript.

\section{Evaluation}

\subsection{Probing} 
We test whether a model anticipates when a backchannel is appropriate by force-decoding annotated human conversations and training a probe to predict from the hidden state at frame $t$ whether a backchannel onset occurs at $t+1$. We probe both the base models and their counterparts fine-tuned with the backchannel head (-BC): since the backbone is fine-tuned jointly with the head (\cref{sec:exp}), training can change the hidden states, and the comparison tests whether it makes upcoming onsets easier to predict. A separate probe, rather than the head itself, lets us measure the base models, which have no head, in the same way.

\paragraph{Method.} For each conversation, we force-decode the ground-truth audio of both speakers and the word-aligned text stream, silencing the real backchannels so that predictions can also be tested during and after them, and extract the hidden states of every temporal-transformer layer. Probing all layers ensures a fair comparison with the base models, which may encode backchannel timing best at an intermediate layer, and shows whether training makes the signal available at the last layer, which the head reads at inference. For each layer, we fit an MLP probe with class-balanced weighting on TurnBench and evaluate it on held-out conversations.

\paragraph{Evaluation.} Using the probe, we count how often $p(y_{t}=1\mid h_t)$ crosses a fixed threshold and compare this rate to the human rate. We separately score whether the predicted timing itself is correct against the human annotations, reporting F1.

\definecolor{highlight}{HTML}{71B1D7}

\begin{table*}[ht]
\centering
\small
\begin{tabular}{lccccccccc}
\toprule
& \textbf{Pause (Synth.)} & \textbf{Pause (Candor)} & \multicolumn{3}{c}{\textbf{Backchannel}} & \multicolumn{2}{c}{\textbf{Smooth Turn Taking}} & \multicolumn{2}{c}{\textbf{User Interruption}} \\
\cmidrule(lr){2-2} \cmidrule(lr){3-3} \cmidrule(lr){4-6} \cmidrule(lr){7-8} \cmidrule(lr){9-10}
\textbf{Model} & \textbf{TOR $\downarrow$} & \textbf{TOR $\downarrow$} & \textbf{TOR $\downarrow$} & \textbf{Freq $\uparrow$} & \textbf{JSD $\downarrow$} & \textbf{TOR $\uparrow$} & \textbf{Latency $\downarrow$} & \textbf{TOR $\uparrow$} & \textbf{Latency $\downarrow$} \\
\midrule
Moshi & \cellcolor{highlight!58!white}0.445 & \cellcolor{highlight!42!white}0.528 & \cellcolor{highlight!31!white}0.255 & \cellcolor{highlight!10!white}0.074 & \cellcolor{highlight!10!white}0.824 & \cellcolor{highlight!10!white}0.739 & \cellcolor{highlight!46!white}0.162 & \cellcolor{highlight!10!white}0.920 & \cellcolor{highlight!10!white}1.377 \\
+ RL \cite{ohashi2026multifacetedinteractivityalignmentfullduplex} & \cellcolor{highlight!95!white}0.226 & \cellcolor{highlight!77!white}0.417 & \cellcolor{highlight!95!white}0.091 & \cellcolor{highlight!46!white}0.095 & \cellcolor{highlight!10!white}0.789 & \cellcolor{highlight!95!white}0.966 & \cellcolor{highlight!72!white}0.121 & \cellcolor{highlight!95!white}1.000 & \cellcolor{highlight!10!white}0.461 \\
\midrule
PersonaPlex & \cellcolor{highlight!45!white}0.482 & \cellcolor{highlight!69!white}0.444 & \cellcolor{highlight!73!white}0.182 & \cellcolor{highlight!10!white}0.046 & \cellcolor{highlight!10!white}0.841 & \cellcolor{highlight!95!white}0.958 & \cellcolor{highlight!10!white}0.219 & \cellcolor{highlight!23!white}0.940 & \cellcolor{highlight!60!white}0.271 \\
+ RL \cite{ohashi2026multifacetedinteractivityalignmentfullduplex} & \cellcolor{highlight!95!white}0.328 & \cellcolor{highlight!95!white}0.361 & \cellcolor{highlight!95!white}0.127 & \cellcolor{highlight!95!white}0.122 & \cellcolor{highlight!18!white}0.783 & \cellcolor{highlight!92!white}0.950 & \cellcolor{highlight!95!white}0.079 & \cellcolor{highlight!95!white}1.000 & \cellcolor{highlight!95!white}0.187 \\
+ BC $_{\text{low}}$ (ours) & \cellcolor{highlight!13!white}0.569 & \cellcolor{highlight!13!white}0.620 & \cellcolor{highlight!10!white}0.291 & \cellcolor{highlight!21!white}0.081 & \cellcolor{highlight!22!white}0.780 & \cellcolor{highlight!95!white}0.958 & \cellcolor{highlight!86!white}0.098 & \cellcolor{highlight!10!white}0.935 & \cellcolor{highlight!47!white}0.294 \\
+ BC $_{\text{normal}}$ (ours) & \cellcolor{highlight!10!white}0.577 & \cellcolor{highlight!10!white}0.630 & \cellcolor{highlight!42!white}0.236 & \cellcolor{highlight!34!white}0.088 & \cellcolor{highlight!25!white}0.778 & \cellcolor{highlight!95!white}0.958 & \cellcolor{highlight!86!white}0.098 & \cellcolor{highlight!10!white}0.935 & \cellcolor{highlight!47!white}0.295 \\
+ BC $_{\text{max}}$ (ours) & \cellcolor{highlight!10!white}0.766 & \cellcolor{highlight!10!white}0.773 & \cellcolor{highlight!52!white}0.218 & \cellcolor{highlight!95!white}0.259 & \cellcolor{highlight!95!white}0.672 & \cellcolor{highlight!95!white}0.983 & \cellcolor{highlight!95!white}0.015 & \cellcolor{highlight!37!white}0.945 & \cellcolor{highlight!95!white}0.181 \\
\midrule
F-Actor & \cellcolor{highlight!81!white}0.409 & \cellcolor{highlight!95!white}0.296 & \cellcolor{highlight!10!white}0.673 & \cellcolor{highlight!44!white}0.102 & \cellcolor{highlight!68!white}0.743 & \cellcolor{highlight!10!white}0.513 & \cellcolor{highlight!10!white}0.675 & \cellcolor{highlight!10!white}0.785 & \cellcolor{highlight!10!white}2.008 \\
+ BC $_{\text{low}}$ (ours) & \cellcolor{highlight!39!white}0.533 & \cellcolor{highlight!67!white}0.463 & \cellcolor{highlight!10!white}0.600 & \cellcolor{highlight!66!white}0.118 & \cellcolor{highlight!95!white}0.721 & \cellcolor{highlight!10!white}0.731 & \cellcolor{highlight!70!white}0.124 & \cellcolor{highlight!28!white}0.925 & \cellcolor{highlight!13!white}1.443 \\
+ BC $_{\text{normal}}$ (ours) & \cellcolor{highlight!95!white}0.343 & \cellcolor{highlight!95!white}0.296 & \cellcolor{highlight!10!white}0.618 & \cellcolor{highlight!71!white}0.122 & \cellcolor{highlight!93!white}0.724 & \cellcolor{highlight!10!white}0.504 & \cellcolor{highlight!10!white}0.905 & \cellcolor{highlight!10!white}0.845 & \cellcolor{highlight!10!white}1.613 \\
+ BC $_{\text{max}}$ (ours) & \cellcolor{highlight!10!white}0.650 & \cellcolor{highlight!30!white}0.569 & \cellcolor{highlight!10!white}0.600 & \cellcolor{highlight!95!white}0.151 & \cellcolor{highlight!95!white}0.709 & \cellcolor{highlight!22!white}0.765 & \cellcolor{highlight!24!white}0.187 & \cellcolor{highlight!10!white}0.920 & \cellcolor{highlight!10!white}1.507 \\
\bottomrule
\end{tabular}
\caption{Comparison of Moshi, PersonaPlex, F-Actor, and our BC variants on Full-Duplex-Bench v1.}\vspace{-0.2cm}
\label{tab:fdbench}
\end{table*}

\subsection{Evaluating the Generated Backchannels}
We test whether the trained head produces correctly timed backchannels once it actually drives generation, and whether doing so comes at the cost of the model's general full-duplex conversational abilities. We evaluate our BC-head-augmented models on Full-Duplex-Bench v1 \cite{lin2025fdb_v1} alongside their base models, isolating the effect of the backchannel head. The benchmark reports backchannel-specific metrics alongside metrics for pause handling, turn-taking, and interruption. We use three thresholds: \textit{normal} matches the human TurnBench rate; \textit{low}/\textit{max} yield fewer/more backchannels.



\subsection{Human Evaluation}\label{subsec:humeval_expl}
In addition to the automatic evaluation, we assess whether human listeners judge the models' backchannels as appropriately placed. We randomly select 35 TurnBench samples of 10--20 seconds, each containing one backchannel, and force-decode them with the PersonaPlex-BC model, letting it predict the backchannel. As a baseline, we also silence the real backchannels and reinsert them at random points. Each sample thus has three versions, human, random, and model, each rated 1--5 for appropriate backchanneling by four annotators (12 in total) via the Pearmut platform \cite{zouhar2026pearmuthumanevaluationtranslation}.

\section{Results}\label{sec:results}

We evaluate the backchannel-head mechanism in three stages: probing its hidden states, generation, and human evaluation.

\subsection{Probing}
\paragraph{Backchannel Frequency.}\cref{fig:frequency} shows the probe's predicted backchannel frequency per layer against the human rate. PersonaPlex-BC reaches this human rate when probing the last layer at a well-calibrated threshold around 0.5. F-Actor-BC likewise improves over its own base model at the last layer, but to a smaller degree. This suggests that the backchannel-head mechanism successfully shifts the model's hidden states toward human-like backchannel frequency.

\paragraph{Onset Timing Accuracy.}
F1 measures whether the probe's predicted onset lands at the correct moment rather than merely how often it fires. \cref{fig:f1_probing} shows PersonaPlex-BC with substantially higher F1 than base PersonaPlex across nearly all layers, while F-Actor-BC's large gain over base F-Actor at middle layers narrows to a small improvement by the last layer. F-Actor(-BC) also performs best at low decision thresholds. A tolerance of two frames ($\pm$0.16s) around the annotated onset improves both models' scores.

\subsection{Evaluating the Generated Backchannels}
We now evaluate the backchannels the model generates.

\paragraph{Full-Duplex-Bench.} \Cref{tab:fdbench} compares the base models against our BC-augmented versions and against RL-based backchannel shaping~\cite{ohashi2026multifacetedinteractivityalignmentfullduplex}. We find that BC backchannels more often and with better timing (lower JSD) than the base models. A lower threshold produces more backchannels, as expected (max vs. low). Compared to RL, PersonaPlex-BC$_{\text{max}}$ achieves higher backchannel frequency, better timing (JSD) and faster turn-taking, but higher take-over rates (TOR), especially on pauses. F-Actor's pause TOR at the normal setting, in contrast, improves over its base model and is among the best in the table.

\paragraph{Surface Forms.} Beyond deciding when to backchannel, the model also has to choose what to say. We test this in two ways: via free generation and via generation constrained to the training data's own token paths. Both produce similar lexical distributions, for example roughly doubling the human share of \textit{yeah}. Despite this lexical similarity, listening to the audio, we find the constrained version sounds more natural.

\begin{figure}
    \centering
    \includegraphics[width=0.9\linewidth]{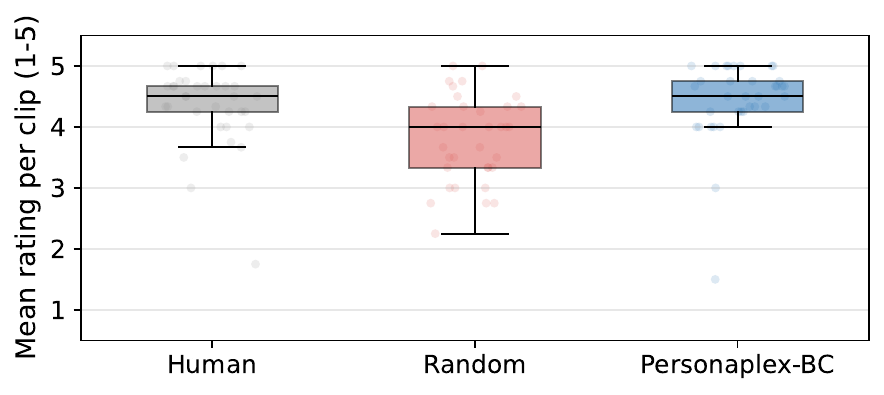}
    \caption{Results of the human evaluation.}
    \label{fig:human_eval_results}\vspace{-0.2cm}
\end{figure}

\subsection{Human Evaluation}
We ask human evaluators to judge backchannel appropriateness (\cref{fig:human_eval_results}): Human recordings score 4.34/5, PersonaPlex-BC 4.37/5, and the random baseline 3.77/5, with both human and PersonaPlex-BC rated significantly higher than random ($p<0.0001$) but not significantly different from each other ($p=0.47$). Inter-annotator agreement (Krippendorff's $\alpha$, interval metric) is $0.42$, moderate and consistent with the inherent subjectivity of naturalness judgments.

\section{Discussion and Conclusion}\label{sec:concl}
We show that backchannel timing can be learned as a light\-weight, pluggable signal, rather than left to emerge implicitly from conversational training. Across two architectures, this explicit control improves backchannel frequency and timing, and in a human evaluation its backchannels are judged on par with human ones. We release our models and code  to make this control mechanism easy to build upon.

\clearpage
\section{Acknowledgments}
This work was supported by JSALT 2026 at Johns Hopkins University with funds from NSF CCRI Grant No. 2120435, Google DeepMind, JHU HLTCOE, JHU AI2AI and ACL, and has received funding from the European Union’s Horizon research and innovation programme under grant agreement No 101135798, project Meetween (My Personal AI Mediator for Virtual MEETtings BetWEEN People). 

Generative AI tools were used to assist with editing and grammar checking of the manuscript, as well as for coding and plotting. All scientific content, analyses, and conclusions were developed and verified by the authors.

\patchcmd{\thebibliography}
  {\advance\leftmargin\labelsep}
  {\setlength{\itemsep}{1pt plus 0.2pt}
   \setlength{\parsep}{1pt}
   \setlength{\parskip}{1pt}
   \advance\leftmargin\labelsep}
  {}
  {}

\bibliographystyle{IEEEtran}
\bibliography{refs}

\end{document}